\documentclass[conference]{IEEEtran}
\IEEEoverridecommandlockouts
\usepackage{cite}
\usepackage{amsmath,amssymb,amsfonts}
\usepackage{graphicx}
\usepackage{textcomp}
\usepackage{xcolor}
\def\BibTeX{{\rm B\kern-.05em{\sc i\kern-.025em b}\kern-.08em
    T\kern-.1667em\lower.7ex\hbox{E}\kern-.125emX}}
    
\usepackage{algorithm}

\usepackage{algpseudocode}

\usepackage{enumitem}
\usepackage{tikz}
\usetikzlibrary{arrows.meta,positioning}

\usepackage{algorithm}
\usepackage{algpseudocode}

\usepackage{enumitem}

\usepackage{hyperref}
\hypersetup{
    colorlinks=true,
    linkcolor=blue,
    filecolor=blue,      
    urlcolor=black,
    citecolor=black,
    pdftitle={Overleaf Example},
    pdfpagemode=FullScreen,
    }

\usepackage{tikz}
\usepackage{textcomp}
\newcommand\copyrighttext{%
  \footnotesize \textcopyright 2026 IEEE. Personal use of this material is permitted.  Permission from IEEE must be obtained for all other uses, in any current or future media, including reprinting/republishing this material for advertising or promotional purposes, creating new collective works, for resale or redistribution to servers or lists, or reuse of any copyrighted component of this work in other works. 

  Accepted for publication in FACCT 2026 Proceedings, Cyber-Physical Systems and Internet-of-Things Week.}
\newcommand{\copyrightnotice}{%
\begin{tikzpicture}[remember picture,overlay]
\node[anchor=south,yshift=10pt] at (current page.south) {\fbox{\parbox{\dimexpr\textwidth-\fboxsep-\fboxrule\relax}{\copyrighttext}}};
\end{tikzpicture}%
}

\begin{document}

\title{Compliance-by-Construction Argument Graphs: Using Generative AI to Produce Evidence-Linked Formal Arguments for Certification-Grade Accountability
}

\author{\IEEEauthorblockN{{Mahyar Tourchi Moghaddam}\\
mtmo@mmmi.sdu.dk\\
\IEEEauthorblockA{\textit{SDU Software Engineering} \\
Odense, Denmark}}}


\maketitle
\copyrightnotice

\begin{abstract}
High-stakes decision systems increasingly require structured justification, traceability, and auditability to ensure accountability and regulatory compliance. Formal arguments commonly used in the certification of safety-critical systems provide a mechanism for structuring claims, reasoning, and evidence in a verifiable manner. At the same time, generative artificial intelligence systems are increasingly integrated into decision-support workflows, assisting with drafting explanations, summarizing evidence, and generating recommendations. However, current deployments often rely on language models as loosely constrained assistants, which introduces risks such as hallucinated reasoning, unsupported claims, and weak traceability.
This paper proposes a compliance-by-construction architecture that integrates Generative AI (GenAI) with structured formal argument representations. The approach treats each AI-assisted step as a claim that must be supported by verifiable evidence and validated against explicit reasoning constraints before it becomes part of an official decision record. The architecture combines four components: {\em i)} a typed Argument Graph representation inspired by assurance-case methods, {\em ii)} retrieval-augmented generation (RAG) to draft argument fragments grounded in authoritative evidence, {\em iii)} a reasoning and validation kernel enforcing completeness and admissibility constraints, and {\em iv)} a provenance ledger aligned with the W3C PROV standard to support auditability.
We present a system design and an evaluation strategy based on enforceable invariants and worked examples. The analysis suggests that deterministic validation rules can prevent unsupported claims from entering the decision record while allowing GenAI to accelerate argument construction. The approach demonstrates how generative models can assist the production of formal arguments while preserving certification-grade accountability, traceability, and human oversight.
\end{abstract}


\section{Introduction}

Formal arguments are central to certification in safety- and security-critical cyber-physical systems (CPS): they organize heterogeneous analyses, surface assumptions, and justify that a system meets requirements in context.
In parallel, high-risk AI deployments in public and regulated decision-making are increasingly subject to certification-like scrutiny, with hard regulatory obligations, documentation, log retention, and expectations of human oversight, making it natural to treat each AI-assisted decision step as a claim that must be discharged by structured argument and evidence \cite{act2024regulation}.

Yet, many real-world deployments treat language models as “helpful assistants” that draft text, summarize, or suggest decisions without enforcing formal completeness, admissibility, or traceability. Current case-management approaches often lack enforceable reasoning and full traceability, thereby amplifying automation bias, leading to procedural errors, and undermining compliance \cite{alon2023human, kang2020algorithmic}. 
These socio-technical failures are compounded by known limitations of large language models (LLMs), especially hallucination (plausible but unsupported outputs) and unreliable uncertainty communication, which are particularly harmful in high-stakes domains such as law, healthcare, and government benefits \cite{huang2025survey}.

Existing literature provides key building blocks but leaves a gap at their intersection. Assurance cases and notations such as Goal Structuring Notation (GSN) formalize “claims-arguments-evidence” structures widely used in safety contexts \cite{GSN2011}. Computational argumentation (e.g., Toulmin-style layouts and Dung-style abstract/structured argumentation) provides representations for structured reasoning, attacks, and assumptions \cite{toulmin2003uses}. 
Meanwhile, retrieval-augmented generation (RAG) can ground LLM outputs in retrieved sources, yet grounding alone does not guarantee that outputs meet domain-specific legal/procedural obligations or that every claim is properly discharged \cite{lewis2020retrieval}. 
 Finally, accountability research emphasizes end-to-end auditability and “reviewability” across the full decision pipeline, not only model-level properties \cite{raji2020closing}. 
 What remains underdeveloped is a compliance-by-construction pipeline in which GenAI accelerates drafting, but formal argument checks and provenance constraints determine what is admissible into the record.

We address this gap by proposing a methodology and reference architecture that turns AI interventions into certification-grade argument artifacts. Technically, we combine {\em i)} a structured Argument Graph (exportable to GSN-like views), {\em ii)} an evidence-gated GenAI drafting workflow (RAG + constrained prompting), {\em iii)} a Reasoning/Validation Kernel that enforces argument constraints and procedural completeness, and {\em iv)} a provenance ledger aligned with W3C PROV and decision-provenance principles \cite{belhajjame2013prov}.

This paper answers the following questions:

\noindent{\bf RQ1:} How can GenAI be used to draft formal(izable) arguments while preventing unsupported claims from entering certification/audit records?

\noindent{\bf RQ2:} Which machine-checkable argument constraints (e.g., evidence linkage, rule coverage, assumption enumeration) provide the highest leverage for “minimal but meaningful” evaluation \cite{cobbe2021reviewable} in high-stakes workflows? 

\noindent{\bf RQ3:} How should provenance and logging be structured so auditors can reconstruct decision pipelines across tools, models, prompts, data, and human interventions?

We contribute {\em i)} a compliance-by-construction argument-generation pipeline; {\em ii)} a concrete Argument Graph schema with validation constraints that supports easy, automatable evaluations; {\em iii)} an integrated provenance model mapping argument steps to W3C PROV; and {\em iv)} worked examples demonstrating how GenAI can assist formal argument creation without weakening accountability.

{\bf Terminology clarification.} We use {\em compliance-by-construction} to denote a design principle in which artifacts are only admitted into a decision record if they satisfy formally specified constraints at construction time, rather than being audited post hoc. {\em Certification-grade} refers to artifacts that are sufficiently structured, traceable, and auditable to support external review in regulated or high-assurance contexts. {\em Grounding} refers to the requirement that generated claims are explicitly linked to identifiable evidence sources, though grounding alone does not guarantee correctness or completeness.

The paper is organized as follows. Section II reviews foundations. Section III presents the methodology. Section IV describes the system architecture. Section V provides guarantees and examples. Section VI discusses implications and limitations, and Section VII concludes.



\section{Background}

Formal arguments are widely used to justify certification claims. In safety and assurance contexts, an “assurance case” generally structures claims, arguments, and evidence; GSN is an established notation that standardizes these elements and their relationships \cite{GSN2011}.
In parallel, foundational argumentation models provide templates for representing and evaluating reasoning. Toulmin’s layout \cite{toulmin2003uses} emphasizes components such as the claim, grounds (evidence), warrants, backing, qualifiers, and rebuttals, supporting a more realistic analysis of arguments than simple premise-conclusion forms. Dung’s abstract argumentation formalizes conflicts (attacks) among arguments and yields semantics for which sets of arguments are acceptable, and structured argumentation frameworks instantiate these ideas with explicit inference trees and defeasible rules \cite{dung1995acceptability}. These representations are directly relevant to certification and audit because they make assumptions explicit and allow systematic checks of completeness and defensibility, not merely readability \cite{GSN2011}.

GenAI introduces a new opportunity: accelerating the authoring of argument fragments, summaries, and evidence mappings. Recent work explores LLMs for assurance-case generation and management, including prompting GPT-4 to generate GSN-compliant safety cases and studying LLMs for automatic assurance-case instantiation from patterns \cite{sivakumar2024prompting}.
More broadly, LLM-focused surveys show rapid growth in argument mining and argument generation, including the use of LLMs to extract argumentative structure and generate counter-arguments \cite{li2025large}.
However, these advances coexist with reliability risks: hallucination remains a central concern in LLM deployments, with surveys documenting a broad taxonomy of hallucination types and mitigation approaches \cite{huang2025survey}. 
 In decision settings, automation bias (overreliance on algorithmic recommendations) has been empirically studied in public-sector decision-making, motivating designs that keep humans meaningfully in control \cite{alon2023human}.

Accountability research suggests that “good explanations” are not sufficient: systems must support review, audit, and contestation throughout the lifecycle and across the full decision pipeline. Internal algorithmic auditing frameworks emphasize process documentation and traceability across development and deployment \cite{raji2020closing}. 
 “Reviewable automated decision-making” advocates record-keeping mechanisms that surface the socio-technical process, rather than focusing solely on model-centric explainability \cite{cobbe2021reviewable}. 
 Decision provenance extends this view, advocating provenance methods to expose decision pipelines and support oversight, audit, compliance, risk mitigation, and user empowerment \cite{singh2018decision}. 
 At a technical standard level, W3C PROV provides a conceptual model for representing provenance, supporting interoperable capture of entities, activities, and agents involved in producing artifacts \cite{belhajjame2013prov}. 

Regulation further pushes high-stakes AI toward certification-like practices. The EU Artificial Intelligence Act (Regulation (EU) 2024/1689) establishes a risk-based framework and imposes requirements for high-risk systems, including transparency, record-keeping/logging, and human oversight \cite{EUAI}. 
 Annex III identifies high-risk use cases including AI used by public authorities to evaluate eligibility for essential public benefits and services \cite{EUAI}. 
 These provisions, together with public-sector transparency initiatives such as the UK Algorithmic Transparency Recording Standard, suggest that “formal arguments + provenance” can become practical compliance artifacts rather than purely academic constructs \cite{uk2023algorithmic}. 


Finally, privacy-preserving learning enables cross-organization training without centralizing raw data. Federated learning (e.g., FedAvg) aggregates local updates \cite{mcmahan2017communication}, while secure aggregation prevents exposure of individual contributions \cite{bonawitz2016practical}. We treat this as an optional subsystem.

\section{Methodology}

\begin{figure}[t]
\centering
\resizebox{\columnwidth}{!}{
\begin{tikzpicture}[
node distance=1.3cm,
box/.style={draw, rectangle, rounded corners, align=center, font=\footnotesize, minimum width=3cm, minimum height=0.8cm},
arrow/.style={->, thick}
]

\node[box] (input) {Case Context\\Inputs};

\node[box, below=of input] (retrieval) {Evidence Retrieval\\(RAG)};

\node[box, below=of retrieval] (genai) {GenAI Drafts Argument\\Claim -- Reason -- Evidence};

\node[box, below=of genai] (validation) {Validation Kernel\\Rules \& Constraints};

\node[box, below left=of validation] (repair) {Repair Loop\\Retrieve / Redraft / Edit};

\node[box, below right=of validation] (persist) {Persist Argument Graph\\+ Provenance};

\node[box, below=of persist] (decision) {Decision Package\\Argument View + Audit Log};

\draw[arrow] (input) -- (retrieval);
\draw[arrow] (retrieval) -- (genai);
\draw[arrow] (genai) -- (validation);

\draw[arrow] (validation) -- node[left]{invalid} (repair);
\draw[arrow] (repair) |- (genai);

\draw[arrow] (validation) -- node[right]{valid} (persist);
\draw[arrow] (persist) -- (decision);

\end{tikzpicture}
}
\caption{Compliance-by-construction pipeline for AI-assisted argument generation. Generative models propose candidate arguments, which are validated against policy constraints before becoming part of the certified record.}
\label{fig:pipeline}
\end{figure}
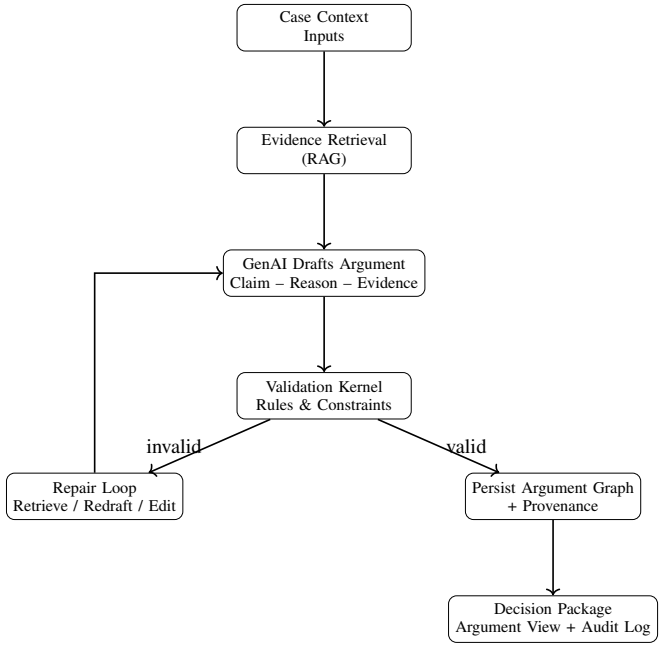

Our methodology treats each AI-assisted workflow as a sequence of certification claims that must be discharged by a structured argument. The core objective is to ensure that only validated argument artifacts can move the workflow forward (“gating”), while GenAI is used for drafting and structuring content, not for deciding admissibility.
Figure~\ref{fig:pipeline} illustrates the compliance-by-construction pipeline in which generative models propose candidate argument fragments while a deterministic validation kernel determines whether they can be admitted into the certified decision record.


We define an Argument Graph (AG) as a typed directed multigraph with node types:

\begin{itemize}[leftmargin=*]
    \item {\em Claim:} a proposition that must be justified (e.g., “Eligibility criterion X is satisfied”).
    \item {\em Rule:} a formalized norm/policy requirement constraining claims (e.g., eligibility, procedural steps).
    \item {\em Evidence:} an admissible source excerpt (documents, records, sensor logs, statutes, policies).
    \item {\em Assumption:} an explicit premise not fully supported by available evidence (must be flagged for review).
    \item {\em Inference/Strategy:} a justification pattern linking subclaims to top-level claims (e.g., GSN strategy nodes).
    \item {\em Uncertainty:} qualifiers or bounds, including confidence and known unknowns (attached to claims, not replacing evidence).
\end{itemize}

This design draws inspiration from assurance-case structures (claims–arguments–evidence) and argumentation models that enforce explicit warrants/assumptions.

\noindent{\bf \em Evidence-gated GenAI drafting.} We use retrieval-augmented generation to provide the model with authoritative snippets and require that each claim node include at least one evidence pointer. RAG provides a practical grounding mechanism beyond parametric memory alone, though it does not eliminate all unsupported generations, hence the need for validation.
However, grounding alone does not guarantee procedural completeness or policy compliance, which motivates the subsequent validation stage.

\noindent{\bf \em Validation kernel (formal checks).} The validator is a deterministic component that enforces constraints before persistence. 
Conceptually, the system follows the principle “GenAI proposes; the validator disposes”: generative models draft candidate argument fragments while deterministic checks determine whether these artifacts satisfy certification constraints. 
This aligns with “by-construction” compliance: correctness is achieved not by hoping generation behaves well, but by rejecting artifacts that violate rules. 

We define a predicate Valid(AG, $\Pi$), where $\Pi$ is the active policy/constraint set (rules, procedures, and governance constraints). Valid holds when all constraints below are satisfied:

\begin{enumerate}[leftmargin=*]
    \item {\bf \em Evidence completeness:} every Claim has $\geq$ 1 Evidence edge (or is explicitly labeled as an Assumption requiring human approval).
    \item  {\bf \em Evidence admissibility:} every Evidence node comes from an authorized source class and within allowed data-governance scope (e.g., approved statutory corpus, approved record types).
    \item {\bf \em Rule coverage:} for each top-level claim class, all required subclaims/procedural steps exist (e.g., mandatory checks).
    \item {\bf \em Non-contradiction (local):} the graph contains no direct conflicts among claims under the chosen consistency rules; conflicts must be represented explicitly as rebuttals/attacks, not silently ignored. 
    \item {\bf \em Provenance completeness:} every AI-generated node must be linked to {\em i)} model identifier/version, {\em ii)} prompt/template version, {\em iii)} retrieval set identifiers, and {\em iv)} human editor actions (if any) via PROV relations. 
\end{enumerate}

Algorithm~\ref{alg:certified_argument} operationalizes the compliance-by-construction principle underlying our approach. The goal is to transform a case description into a structured argument graph whose claims are explicitly supported by admissible evidence and validated against policy constraints before becoming part of the decision record.
The procedure begins by constructing a \emph{case knowledge graph} from the input case $C$. This representation organizes relevant entities, events, and relationships extracted from the case data, providing a structured context for downstream reasoning and evidence retrieval. Using this graph and the active policy constraints $\Pi$, the system retrieves a set of potentially relevant evidence sources $R$. Retrieval is implemented using a retrieval-augmented generation (RAG) mechanism, which identifies authoritative documents or records that may support claims within the argument.

Given the knowledge graph and retrieved evidence, a generative model produces a candidate argument fragment. The output is constrained to a predefined schema so that it can be interpreted as a typed argument graph rather than free-form text. The generated structure is then parsed and normalized to ensure consistency with the argument representation used by the reasoning system.

\begin{algorithm}[t]
{\scriptsize
\caption{BuildCertifiedArgument}
\label{alg:certified_argument}

\begin{algorithmic}[1]

\State \textbf{Input:} Case $C$, policy constraints $\Pi$
\State \textbf{Output:} Certified argument graph $AG$ and provenance record $P$

\State $KG \gets$ BuildCaseKnowledgeGraph($C$)

\State $R \gets$ RetrieveEvidence($KG, \Pi$) \Comment{RAG retrieval set}

\State $draft \gets$ GenAI\_DraftArgument($KG, R, \Pi, schema=AG$)

\State $AG \gets$ ParseAndNormalize($draft$)

\If{Valid($AG,\Pi$)}

    \State $P \gets$ EmitProvenance($AG, KG, R, model\_meta, human\_meta$)

    \State Persist($AG, P$)

    \State \Return $(AG,P)$

\Else

    \State $feedback \gets$ ExplainViolations($AG,\Pi$)

    \State \Return RepairLoop($C,\Pi,feedback$)

\EndIf

\end{algorithmic}}
\end{algorithm}

The resulting argument graph $AG$ is evaluated by the validation kernel. This component performs deterministic checks to verify that the argument satisfies the policy constraints $\Pi$. These checks ensure, for example, that each claim is linked to admissible evidence, required reasoning steps are present, and structural requirements of the argument schema are respected.
If the argument graph satisfies these constraints, the system records provenance metadata describing how the artifact was produced, including the evidence sources, model invocation, and any human interaction involved in the process. The validated argument graph and its provenance record are then persisted as part of the auditable decision package.
If validation fails, the system produces a structured explanation of the violated constraints and initiates a repair loop. This loop may involve re-retrieving evidence, regenerating the argument fragment, or incorporating human corrections before re-evaluating the resulting graph.
The algorithm thus separates generative proposal from formal acceptance: generative models produce candidate arguments, while the validation kernel determines whether they satisfy the constraints required for inclusion in the certified argument graph.

\section{System Design}

\begin{figure}[t]
\centering
\includegraphics[width=.6\columnwidth]{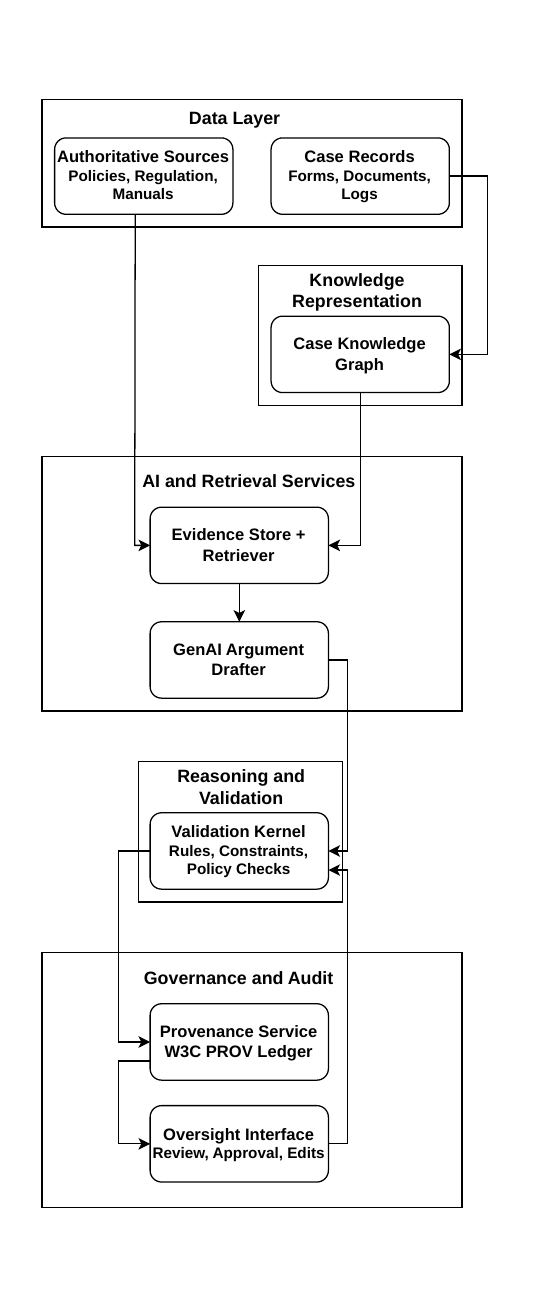}
\caption{Layered architecture for compliance-by-construction argument generation.}
\label{fig:architecture}
\end{figure}

We describe an implementable reference system with six components, designed for certification authorities and auditors to consume outputs as formal arguments rather than opaque model text.
Figure~\ref{fig:architecture} presents the layered architecture implementing this workflow, separating data ingestion, knowledge representation, AI-assisted argument drafting, deterministic validation, and governance components.

\noindent{\bf \em Knowledge representation layer (Case Knowledge Graph).} The system normalizes case context into structured entities (people/orgs/assets), events, and relationships. This design supports “traceability between models and requirements” by enabling stable IDs and links between argument claims and underlying facts. 

\noindent{\bf \em Evidence layer (Retriever + admissibility).} We store retrieved evidence chunks with immutable identifiers, capturing retrieval parameters and access controls. RAG is the main grounding mechanism, based on well-established retrieval-augmented generation formulations. 
 Evidence admissibility rules enforce that only permitted corpora contribute to certification artifacts, mitigating a common failure mode where LLMs cite non-authoritative or fabricated sources. 

\noindent{\bf \em GenAI argument drafter (bounded authoring).} The model is prompted to output only a typed AG structure, not free-form narrative. To reduce unbounded generation, we constrain outputs via JSON schemas / typed templates and require that each claim references evidence IDs from the retrieval set. This step is inspired by the broader research trend of using LLMs to draft assurance case fragments, while acknowledging that generation must be checked. 

\noindent{\bf \em Validation kernel (deterministic).} The kernel is the system’s reliability boundary. It checks: evidence coverage, procedural completeness, and provenance completeness. AI is thus embedded as an infrastructure component operating under explicit reasoning and governance constraints rather than as an opaque assistant.  
 It also aligns with EU AI Act expectations for logging and human oversight for high-risk systems. 

\noindent{\bf \em Provenance service (W3C PROV + decision provenance).} Provenance is emitted both for design-time artifacts (prompt templates, validator policies, model cards) and run-time events (retrieval sets, model calls, human edits, acceptance decisions). We align on W3C PROV to improve interoperability while meeting decision-provenance goals for exposing decision pipelines. 
 This design also builds on the broader ecosystem of PROV-compliant ML provenance tooling as evidence that PROV is practical in modern data/ML stacks. 

\noindent{\bf \em Oversight UI (argument consumption).}
Our UI prioritizes {\em i)} argument navigation (GSN-like view); {\em ii)} inspection of evidence snippets and their source metadata; {\em iii)} explicit display of assumptions and uncertainty qualifiers; and {\em iv)} human override and annotation, addressing risks of automation bias. 

\noindent{\bf \em Privacy-preserving learning fabric.} Where cross-organization model improvement is desired, the system can incorporate federated learning with secure aggregation so that raw case data never leaves local control. 

\section{Architecture Guarantees and Representative Examples}


Since the architecture enforces reasoning constraints at validation time, evaluation focuses on properties that can be verified directly from system design and execution traces rather than predictive performance metrics. The analysis, therefore, examines whether the architecture guarantees key reasoning invariants and whether generated artifacts remain auditable and structurally correct under representative failure scenarios.
The results therefore combine {\em i)} architectural guarantees that hold by construction and {\em ii)} representative examples demonstrating system behavior when generative outputs are incomplete or unsupported.

\subsection{Evidence completeness as an enforceable invariant}

The validation kernel ensures that every accepted argument graph satisfies a strict evidence completeness constraint. Formally, if an argument graph \(AG\) is accepted under policy constraints \(\Pi\), then each claim node must either reference at least one admissible evidence node or be explicitly marked as an assumption requiring human approval. This property is not probabilistic but deterministic: unsupported claims cannot enter the persisted argument record because they are rejected during validation.

This constraint directly addresses a central reliability risk of generative systems: the production of plausible but unsupported reasoning. While generative models may produce statements that appear coherent, such statements are excluded from the certified argument graph unless they are grounded in retrieved evidence or explicitly identified as assumptions.
As a result, hallucinated claims can occur during drafting but cannot propagate into the final decision artifact without review.

\subsection{Audit reconstruction through provenance mapping}

Each accepted argument graph is associated with a provenance record capturing the activities, inputs, and agents involved in its creation. The provenance structure follows the W3C PROV model and records relationships between argument artifacts, evidence sources, model invocations, and human interventions.

This representation enables reconstruction of the reasoning process that produced a decision artifact. For example, auditors can determine:

\begin{itemize}[leftmargin=*]
\item which evidence fragments support a particular claim,
\item which model version produced a draft argument fragment, and
\item which human reviewer approved or modified the resulting argument.
\end{itemize}

Such queries can be answered by traversing provenance relations linking entities, activities, and agents. This capability enables full traceability from a final decision back to the intermediate reasoning steps and data sources that produced it.

\subsection{Procedural completeness and repair loop behavior}

A common failure mode of generative systems occurs when required procedural steps are omitted from a generated explanation or reasoning structure. In workflows governed by policy constraints, certain checks must always be present, for instance verifying identity, evaluating eligibility criteria, or documenting the rationale for a decision.

When the generative component produces an argument graph missing one of these required elements, the validation kernel detects the violation and rejects the artifact. The kernel then returns a structured explanation describing the missing constraint. The system uses this feedback to trigger a repair loop in which the argument fragment is regenerated, revised, or manually corrected before revalidation.

This mechanism ensures that incomplete reasoning structures cannot advance through the workflow. Instead, the system iteratively refines the argument until all structural and evidentiary requirements are satisfied.

\subsection{Example argument structure}

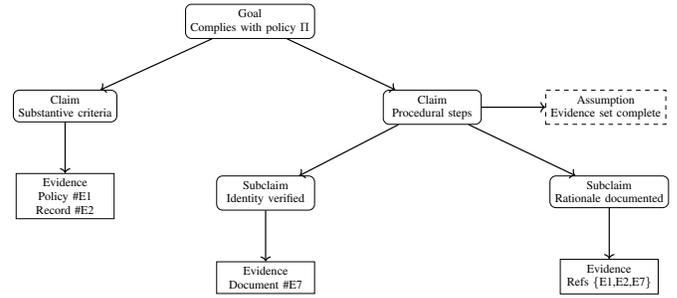
\begin{figure}[t]
\centering
\resizebox{\columnwidth}{!}{
\begin{tikzpicture}[
node distance=1.2cm and 1.6cm,
box/.style={draw, rectangle, rounded corners, align=center, font=\footnotesize, minimum width=2.3cm, minimum height=0.7cm},
evidence/.style={draw, rectangle, align=center, font=\footnotesize, minimum width=2.3cm, minimum height=0.7cm},
assumption/.style={draw, rectangle, dashed, align=center, font=\footnotesize, minimum width=2.3cm, minimum height=0.7cm},
arrow/.style={->, thick}
]

\node[box] (goal) {Goal\\Complies with policy $\Pi$};

\node[box, below left=of goal] (claim1) {Claim\\Substantive criteria};
\node[box, below right=of goal] (claim2) {Claim\\Procedural steps};

\node[evidence, below=of claim1] (e1) {Evidence\\Policy \#E1\\Record \#E2};

\node[box, below left=of claim2] (sub1) {Subclaim\\Identity verified};
\node[box, below right=of claim2] (sub2) {Subclaim\\Rationale documented};

\node[evidence, below=of sub1] (e2) {Evidence\\Document \#E7};
\node[evidence, below=of sub2] (e3) {Evidence\\Refs \{E1,E2,E7\}};

\node[assumption, right=1.5cm of claim2] (ass) {Assumption\\Evidence set complete};

\draw[arrow] (goal) -- (claim1);
\draw[arrow] (goal) -- (claim2);

\draw[arrow] (claim1) -- (e1);

\draw[arrow] (claim2) -- (sub1);
\draw[arrow] (claim2) -- (sub2);

\draw[arrow] (sub1) -- (e2);
\draw[arrow] (sub2) -- (e3);

\draw[arrow] (claim2) -- (ass);

\end{tikzpicture}
}
\caption{Example argument graph fragment showing hierarchical claims, evidence links, and explicit assumptions.}
\label{fig:argument-example}
\end{figure}

Figure~\ref{fig:argument-example} illustrates a simplified fragment of an argument graph generated by the system. The structure follows a hierarchical pattern similar to assurance-case representations, where a top-level claim is decomposed into subclaims supported by evidence and reasoning strategies.

The example illustrates how generative models can assist with drafting the structure and wording of argument elements, while the validation kernel ensures that all claims remain grounded in evidence and that assumptions are explicitly identified.

\section{Discussion}

%
The results highlight several design implications for integrating generative models into systems that require verifiable reasoning and institutional accountability.
First, generative models are most effective when used as \emph{structured drafting assistants rather than autonomous reasoning agents}. Language models are capable of rapidly producing candidate argument fragments, but their outputs remain probabilistic and may omit necessary reasoning steps. Placing a deterministic validation layer between generation and persistence ensures that such outputs are treated as proposals rather than authoritative conclusions.
Second, evaluating systems of this kind requires shifting focus from predictive accuracy toward \emph{structural guarantees}. Instead of attempting to prove that generated reasoning is always correct, the evaluation demonstrates that unsupported claims cannot be accepted into the final artifact. This property is particularly important in contexts where explanations must remain verifiable and defensible.
Third, provenance plays a central role in maintaining accountability. By capturing the full sequence of activities involved in producing an argument graph, including evidence retrieval, model invocation, validation, and human edits, the system enables auditors to reconstruct how a decision artifact was produced. This capability goes beyond traditional model explanations by exposing the entire reasoning pipeline rather than only the model output.

Finally, the approach illustrates how GenAI can be integrated into decision-support workflows without compromising transparency. Instead of replacing structured reasoning processes, generative models accelerate the construction of formal arguments while deterministic validation preserves their integrity.
Despite these advantages, several limitations remain. Encoding policies as machine-checkable constraints requires formalization decisions that may not capture all contextual nuances. In addition, while structural validation ensures evidence linkage and procedural completeness, it does not guarantee that the underlying policy rules themselves are unbiased or normatively correct. Addressing these issues requires complementary governance mechanisms, including auditing processes, documentation standards, and oversight practices.
Future work should therefore explore richer argumentation models capable of representing conflicting claims, uncertainty propagation, and adversarial reasoning. Such extensions would enable the system to handle contested decisions and complex policy interpretations more effectively.

\section{Conclusion}

This paper presented a compliance-by-construction architecture for generating evidence-linked argument structures using GenAI and deterministic validation. The approach separates drafting from acceptance: generative models produce candidate artifacts, while a validation kernel ensures that only those that satisfy formal constraints are added to the certified record.
The system combines three core elements: a structured argument graph, a validation kernel enforcing evidence linkage and procedural completeness, and a provenance model enabling full reconstruction of the reasoning pipeline. Together, these support integration of generative models into high-stakes workflows while maintaining auditability and traceability.
The analysis demonstrates enforceable invariants that prevent unsupported claims from propagating into final artifacts. More broadly, the work shows that treating model outputs as candidate artifacts subject to deterministic verification enables the benefits of generative models while preserving accountability in regulated environments.

\bibliographystyle{IEEEtran}
\bibliography{bib}

@ArtifactSoftware{R,
    title = {R: A Language and Environment for Statistical Computing},
    author = {{R Core Team}},
    organization = {R Foundation for Statistical Computing},
    address = {Vienna, Austria},
    year = {2019},
    url = {https://www.R-project.org/},
}

@article{act2024regulation,
  title={Regulation (eu) 2024/2847 of the european parliament and of the council, Act Resilience},
  author={European Union, R},
  journal={Regulation (eu)},
  year={2024}
}

@article{EUAI,
  title={Regulation (eu) 2024/1689 of the european parliament and of the council},
  author={Act, AI},
  journal={Regulation (eu)},
  year={2024}
}

@misc{uk2023algorithmic,
  title={Algorithmic transparency recording standard hub},
  author={UK, GOV},
journal={www.gov.uk},
  year={2023}
}

@article{huang2025survey,
  title={A survey on hallucination in large language models: Principles, taxonomy, challenges, and open questions},
  author={Huang, Lei and Yu, Weijiang and Ma, Weitao and Zhong, Weihong and Feng, Zhangyin and Wang, Haotian and Chen, Qianglong and Peng, Weihua and Feng, Xiaocheng and Qin, Bing and others},
  journal={ACM Transactions on Information Systems},
  volume={43},
  number={2},
  pages={1--55},
  year={2025},
  publisher={ACM New York, NY}
}

@article{alon2023human,
  title={Human--AI interactions in public sector decision making:“automation bias” and “selective adherence” to algorithmic advice},
  author={Alon-Barkat, Saar and Busuioc, Madalina},
  journal={Journal of Public Administration Research and Theory},
  volume={33},
  number={1},
  pages={153--169},
  year={2023},
  publisher={Oxford University Press US}
}

@inproceedings{kang2020algorithmic,
  title={Algorithmic accountability in public administration: The GDPR paradox},
  author={Kang, Sunny Seon},
  booktitle={Proceedings of the 2020 Conference on Fairness, Accountability, and Transparency},
  pages={32--32},
  year={2020}
}

@techreport{GSN2011,
  title        = {GSN Community Standard Version 1},
  author       = {{Goal Structuring Notation Working Group}},
  institution  = {Origin Consulting (York) Ltd.},
  year         = {2011},
  month        = nov,
  url          = {https://www.faa.gov/about/office_org/headquarters_offices/ang/redac/redac-sas-201503-gsn-community-standard-v1.pdf},
  note         = {Accessed: 2026-03-08}
}

@book{toulmin2003uses,
  title={The uses of argument},
  author={Toulmin, Stephen E},
  year={2003},
  publisher={Cambridge university press}
}

@article{lewis2020retrieval,
  title={Retrieval-augmented generation for knowledge-intensive nlp tasks},
  author={Lewis, Patrick and Perez, Ethan and Piktus, Aleksandra and Petroni, Fabio and Karpukhin, Vladimir and Goyal, Naman and K{\"u}ttler, Heinrich and Lewis, Mike and Yih, Wen-tau and Rockt{\"a}schel, Tim and others},
  journal={Advances in neural information processing systems},
  volume={33},
  pages={9459--9474},
  year={2020}
}

@inproceedings{raji2020closing,
  title={Closing the AI accountability gap: Defining an end-to-end framework for internal algorithmic auditing},
  author={Raji, Inioluwa Deborah and Smart, Andrew and White, Rebecca N and Mitchell, Margaret and Gebru, Timnit and Hutchinson, Ben and Smith-Loud, Jamila and Theron, Daniel and Barnes, Parker},
  booktitle={Proceedings of the 2020 conference on fairness, accountability, and transparency},
  pages={33--44},
  year={2020}
}

@article{belhajjame2013prov,
  title={Prov-dm: The prov data model},
  author={Belhajjame, Khalid and B’Far, Reza and Cheney, James and Coppens, Sam and Cresswell, Stephen and Gil, Yolanda and Groth, Paul and Klyne, Graham and Lebo, Timothy and McCusker, Jim and others},
  journal={W3C Recommendation},
  volume={14},
  pages={15--16},
  year={2013},
  publisher={World Wide Web Consortium (W3C)}
}

@inproceedings{cobbe2021reviewable,
  title={Reviewable automated decision-making: A framework for accountable algorithmic systems},
  author={Cobbe, Jennifer and Lee, Michelle Seng Ah and Singh, Jatinder},
  booktitle={Proceedings of the 2021 ACM conference on fairness, accountability, and transparency},
  pages={598--609},
  year={2021}
}

@article{dung1995acceptability,
  title={On the acceptability of arguments and its fundamental role in nonmonotonic reasoning, logic programming and n-person games},
  author={Dung, Phan Minh},
  journal={Artificial intelligence},
  volume={77},
  number={2},
  pages={321--357},
  year={1995},
  publisher={Elsevier}
}

@article{sivakumar2024prompting,
  title={Prompting GPT--4 to support automatic safety case generation},
  author={Sivakumar, Mithila and Belle, Alvine B and Shan, Jinjun and Shahandashti, Kimya Khakzad},
  journal={Expert Systems with Applications},
  volume={255},
  pages={124653},
  year={2024},
  publisher={Elsevier}
}

@article{li2025large,
  title={Large language models in argument mining: A survey},
  author={Li, Hao and Schlegel, Viktor and Sun, Yizheng and Batista-Navarro, Riza and Nenadic, Goran},
  journal={arXiv preprint arXiv:2506.16383},
  year={2025}
}

@article{singh2018decision,
  title={Decision provenance: Harnessing data flow for accountable systems},
  author={Singh, Jatinder and Cobbe, Jennifer and Norval, Chris},
  journal={IEEE Access},
  volume={7},
  pages={6562--6574},
  year={2018},
  publisher={IEEE}
}

@inproceedings{mcmahan2017communication,
  title={Communication-efficient learning of deep networks from decentralized data},
  author={McMahan, Brendan and Moore, Eider and Ramage, Daniel and Hampson, Seth and y Arcas, Blaise Aguera},
  booktitle={Artificial intelligence and statistics},
  pages={1273--1282},
  year={2017},
  organization={Pmlr}
}

@article{bonawitz2016practical,
  title={Practical secure aggregation for federated learning on user-held data},
  author={Bonawitz, Keith and Ivanov, Vladimir and Kreuter, Ben and Marcedone, Antonio and McMahan, H Brendan and Patel, Sarvar and Ramage, Daniel and Segal, Aaron and Seth, Karn},
  journal={arXiv preprint arXiv:1611.04482},
  year={2016}
}

\end{document}